\documentclass[a4paper]{article}

\usepackage{times}
\usepackage{latexsym}

\usepackage{amsmath}
\usepackage{graphicx}
\usepackage{dirtytalk}
\usepackage{booktabs}
\usepackage[breaklinks=true]{hyperref}
\usepackage{breakcites}
\usepackage{microtype}

\usepackage[square,numbers]{natbib}
\numberwithin{equation}{subsection}

\title{Parameter-Efficient Neural Question Answering Models via Graph-Enriched Document Representations}
\author{
    Louis Castricato \\
    Georgia Tech\\
    Georgia \\
    USA \\
    \texttt{lcastricato3@gatech.edu} 
    \and
    Stephen Fitz \\
    Keio University \\
    Tokyo \\
    Japan \\
    \texttt{stephenf@keio.jp} 
    \and
    Won Young Shin \\
    University of Toronto \\
    Ontario \\
    Canada \\
}

\date{}

\begin{document}
\maketitle

\begin{abstract}
As the computational footprint of modern NLP systems grows, it becomes increasingly important to arrive at more efficient models.
We show that by employing graph convolutional document representation, we can
arrive at a question answering system that performs comparably to, and in some
cases exceeds the SOTA solutions, while using less than 5\% of their resources
in terms of trainable parameters. 
As it currently stands, a major issue in applying GCNs to NLP is document representation.
In this paper, we show that a GCN enriched document representation greatly improves the results seen in HotPotQA, even when using a trivial topology.
Our model (gQA), performs admirably when compared to the current SOTA, and requires little to no preprocessing.
In \cite{Shao20}, the authors suggest that
graph networks are not necessary for good performance in multi-hop QA. In this paper, we suggest that large language models are not necessary for
good performance by showing a na\"{i}ve implementation of a GCN performs
comparably to SoTA models based on pretrained language models.
\end{abstract}

\section{Introduction}

In the past decade, Deep Learning (DL) approaches have dominated the field of
Artificial Intelligence (AI) across a wide a array of domains.
Recent developments in neural information processing methods sparked a
revolution in Natural Language Processing (NLP), which resembles advances made
in Computer Vision (CV) at the beginning of this new chapter of AI development.
This progress was made possible through increasingly more advanced
representation methods of natural language inputs.
Initially shallow pre-training of early model layers became standard in NLP
research through methods such as word2vec \cite{mikolov2013distributed}.
Subsequent progress followed trends similar to those in CV, which naturally led
to pre-training of multiple layers of abstraction.
These advancements resulted in progressively deeper hierarchical language representations, such as those derived using self-attention mechanisms in transformer-based architectures \cite{vaswani2017attention}.
Currently SOTA NLP systems use representations derived from pre-training of
entire language models on large quantities of raw text, and often involve
billions of parameters.
Neural language modelling methods came to prominence in recent years due to the
development of techniques such as BERT \cite{devlin2018bert}.
The success of neural network based Machine Learning (ML) models, especially
those involving very deep architectures, can be attributed to their ability to
derive informative embeddings of raw data into submanifolds of real vector
spaces.
The common idea behind these developments is that we can learn syntax and
semantics of natural languages by training a DL model in a self-supervised
fashion on a corpus of raw text.
The general language representations of words in vector spaces induced by modern
neural NLP models can be transferred to other domain-specific architectures and
further tuned for downstream tasks.
One could argue that this transfer learning benefit, as well as straightforward
applicability to cascaded and multi-task learning, is the most exciting recent
development in representation learning research in the context of natural
language data.

However, in many applications of practical interest in the area of NLP, a need
arises to represent language on a supra-lexical level.
In this paper we concern ourselves with the evaluation of modern representation methods on document level.
Some canonical tasks requiring such coarser embedding methods include document
classification, question answering, and summarization.

Traditionally, document embedding methods combined token-level embeddings by
mapping them to a single vector through a variety of models (e.g. vector
averaging and word2vec inspired methods \cite{le2014distributed, 
kenter2016siamese,
hill2016learning,arora2016simple, pagliardini2017unsupervised, kiros2015skip, hill2016learning, chen2017efficient, logeswaran2018efficient}, attention mechanisms
\cite{vaswani2017attention, devlin2018bert, reimers2019sentence, radford2018improving, radford2019language}, convolutional
architectures \cite{liu2017learning}, or gated recurrent units
\cite{hochreiter1997long, cho2014learning}).

 \subsection{Arbitrary Topologies}

 A \textbf{Document Topology} is the adjacency matrix that determines
 relationships between parts of the document (usually sentences or paragraphs).
 For instance, consider a news article like those
 found in CNN/Dailymail \cite{See17}. Within this dataset, the document topology
 resembles a sparse diagonal matrix, implying that the sentences are meant to be read
 sequentially in the order they appear. A family of neural embedding models
taking graph topology into  account can be grouped under the label of graph
neural networks (GNN). In recent years multiple instantiations of this idea
appeared in literature \cite{bruna2013spectral, henaff2015deep,
duvenaud2015convolutional, li2015gated, 
defferrard2016convolutional, kipf2016semisupervised}.

 As an example of a downstream task which benefits from GCN encoding consider
 question answering on the HotPotQA Dataset, \cite{Yang18}. This is a multi
 document question answering dataset where documents may or may not be related to
 the question. Additionally, some questions may require inferring from "multiple
 hops" within documents to reach a conclusive answer.

 \subsection{Advantages to non-na\"{i}ve representation.}

A common theme in contemporary multi-doc QA can be defined as figuring out the
topology of a set of inference hops. Consider approaches like HGN and DFQN
\cite{Fang19, Xiao19} where the topology is constructed by following
relevant corefences to bridge paragraphs and documents together. While they both
report above baseline performance, the preprocessing step is nontrivial as it
requires ranking paragraphs using large pretrained language models. By
comparison, we require no such step.

Methods closer to ours, like QFE and KGNN \cite{Nishida19, Ye19}  which
possess identical decoders and encoders, albeit they do not encode sentences
disjointly, similarly use multihop reasoning and do not require large language
models. Thus these four will be our main points of comparison.

In the following section, we introduce gQA our test model for multi-document
question answering. The goal of this project wasn't to beat SOTA results on QA,
but rather to evaluate the performance of naive represnetations for multi-doc
QA. 

However, to our surprise, introducing graphical representation on document
topology to a baseline QA approach greatly improved performance, resulting in a
model that performs competitively with SOTA systems, while avoiding costly
pre-training of over-parametrized language models.
This led us to the realization that a graph based representation can produce more 
sustainable multi-doc QA systems by significantly reducing model complexity, and hence lowering
the cost of model training.

 \section{Model}

 \begin{figure*}[t]
 \hspace{-2.5cm} \includegraphics[width=1.4\textwidth]{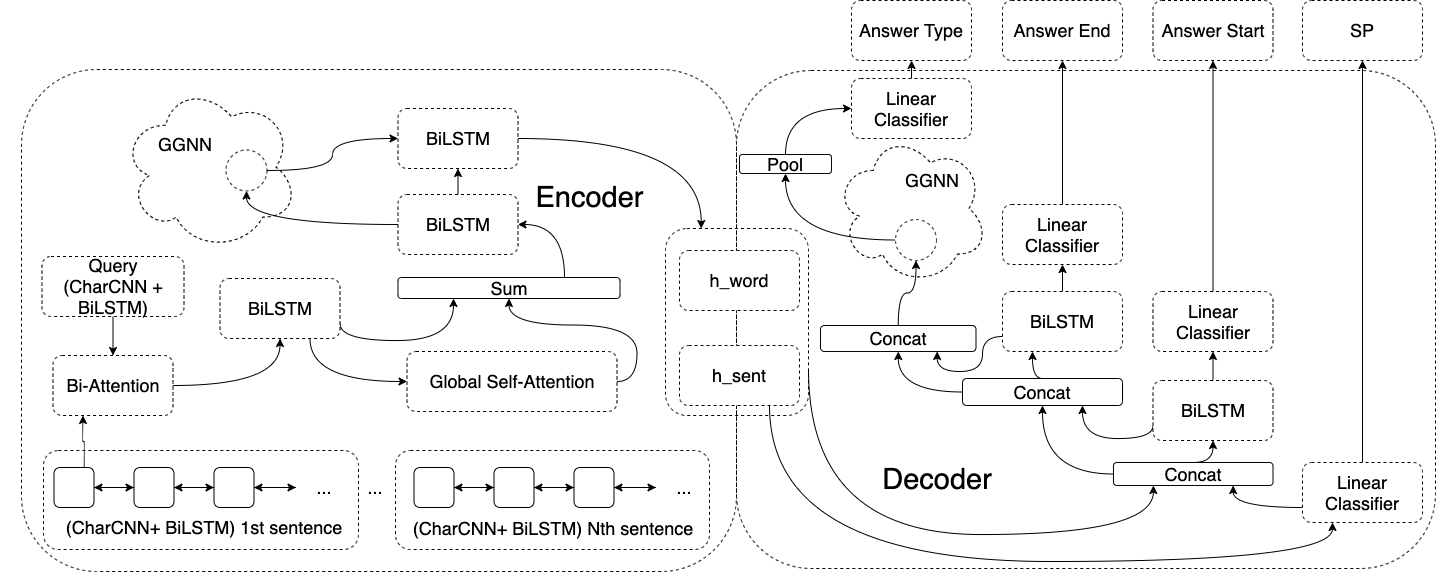}\hspace*{\fill}

 \begin{center}

 \caption{Diagram of the gQA model. Sentences are encoded disjointly, and then
enriched with an embedding of the question. Global self attention is then
applied to all documents at a word level, which is then summed to the query enrichment of the
words. A third BiLSTM is intialized to these summed embeddings. The
hidden state of this BiLSTM becomes the sentence embedding. This
sentence embedding is fed to a GGNN with our document topology. A fourth BiLSTM
is initialized to the output of the GGNN and fed the output of the third BiLSTM
as input. The (output, hidden state) tuple is then fed to the decoder, which is a series of
BiLSTM+Linear Classifier, followed by concating onto the word and sentence
embeddings. Finally, answer type is determined by feeding concatenaded
sentence embeddings to a GGNN, pooling the output via a weighted average, and
then feeding the pooled output to a linear classifier.}
 \end{center}
 \end{figure*}

 In the design of gQA, we aimed for a model that would work over
 arbitrary input topologies for multi-document question answering.

 \subsection{Architecture}

\subsection{Separate Encoding of Sentences}

 Inspired by \cite{Qian18}, we introduce individual sentence representations.
 Given that sentences in Wikipedia, and
 particularly HotPotQA, can be viewed as "supporting facts" for questions rather
 than outlining some long detailed chronology it made sense to consider these
 sentences as individual nodes in a larger concept graph.

 In GraphIE \cite{Qian18}, an identical approach is utilized by considering twitter
 accounts as individual nodes and performing NER over the entire graph.

 \subsection{Encoder}
By contrast, when compared to GraphIE \cite{Qian18}, we consider the GNN as well as
the"decoder" BiLSTM as part of our encoder. That is to say, we encode the
sentences  disjointly, perform inter-sentence enrichment via the Gated Graph
Neural Network (GGNN) \cite{Li2017}, and then intra-sentence enrichment by
initializing a BiLSTM to the enriched sentence
embeddings and feeding the word embeddings as input. GNN's or GCN's are models
that capture graphical information via message passing between the nodes of a
graph. Gated GNNs incorporate gating mechanisms that can learn to allow, block,
or forget information that attempts to pass through the cells in GGNNs.

The propagation and output model for the GGNN is described by the following set of equations.
\footnote{A soft attention mechanism is applied with
$\sigma(i(h_v^{\text{T}},x_v))$, which in this case is the sentence level
attention. $h_v^\text{(T)}$ and $x_v$ are concatenated and input into neural
networks $i()$ and $j()$, which outputs real-valued vectors. The tanh$(\cdot)$ function
can be replaced with an identity function.}
\begin{align*}
 h_v^{\text{(1)}}&=[x_v^\top,0]^\top\\
 a_v^{\text{(t)}}&=A_v^\top [h_1^{\text{(t-1)}\top} ...
 h_{\text{N}}^{\text{(t-1)}^\top}]^\top + b\\
 z_v^t &= \sigma(W^z a_v^{\text{(t)}} + U^z h_v^{\text{(t-1)}})\\
 r_v^t &= \sigma(W^r a_v^{\text{(t)}} + U^r h_v^{\text{(t-1)}})\\
 \widetilde{h_v^{\text{(t)}}}&=\text{tanh}(W a_v^{\text{(t)}} + U(r_v^t \odot h_v^{\text{(t-1)}}))\\
 h_v^{\text{(t)}}&=(1-z_v^t)\odot h_v^{\text{(t-1)}} + z_v^t \odot
\widetilde{h_v^{\text{(t)}}}
 \end{align*} 

In the formulae above, $A$ denotes the adjacency matrix of our GGNN, $h_v^{(t)}$
denotes the hidden state of vertex $v$ at time $t$, and $x_v$ denotes the vector
input to vertex $v$. Finally, $W,U,b$ are all learnable parameters.

 In order to ablate the contribution of document representation, we wanted this
to be a na\"{i}ve implementation of GGNNs on top of the baseline  HotPotQA. For
this purpose, we introduce a Bi-Attention and Self-Attention layer preceding the
GNN. Both Bi-Attention and Self-Attention are done over the entire set of
documents. Bi-Attention is conditioned on the question.  Excluding the GNN, our
encoder is identical to that presented in the baseline of  HotPotQA \cite{Yang18}.

 For the first layer of our encoder, we used a BiLSTM + CharCNN, with GloVe
 embeddings.

\subsection{Decoder}

 The head utilized is nearly identical to that presented in HotPotQA's baseline, with
 the exclusion of how \say{answer\_type} is computed.

 \subsubsection{Supporting Sentences}

 Given the output of the post-GGNN BiLSTM encoder, the sentence embeddings are
fed to a linear classification layer in order to determine the supporting
sentences.

 The contextualized word embeddings and contextualized sentence embeddings for
 the supporting sentences are then concatenated onto $h_{\text{word}}$ and
 $h_{\text{sent}}$ respectively.

 \subsubsection{Answer Start}

 Given $(h_{\text{word}}, h_{\text{sent}})$ from the supporting sentence
 computation, a sentence level BiLSTM is initialized to $h_{\text{sent}}$ and
 given $h_{\text{word}}$ as input.

 The word level output of this BiLSTM is fed to a linear classification layer,
 and the BiLSTM output is concatenated onto the latent word and
 sentence embeddings.

 \subsubsection{Answer End}

 Answer end is computed identically to answer start, including the output being
 concatenated onto the latent embeddings.

 \subsubsection{Answer Type}

A GGNN is fed $h_{\text{sent}}$ as input, which is then fed to a sentence level
attention pooling layer and  eventually another linear classification layer.

 \section{Preprocessing}
We define the adjacency matrix used in our GGNNs by the following formula
\[M = \frac{1}{2}(B + (\lambda)(S - I) + L)^T(B + (\lambda)(S - I) + L)\]
 Where $B$ is a rotated diagonal matrix, $\lambda(S-I)$ is the
regularization parameter \footnote{$S - I$ refers to the cosine similarity
matrix with a zero'ed diagonal. During training, $\lambda = 0.5$ but during
testing $\lambda=0.05$. $\lambda$ represents the only hyperparameter search
conducted.}, and $L$ is an adj. matrix that connects the last
sentence of every document to the first sentence of every other.
\section{Results}

Results of our sustainable QA model are compared to top models at the time of writing.
Our model performs at SoTA level while utilizing less than 5\%, and near SoTA
(still beating some of the top models), with less than 1\% of trainable parameters.
 gQA-XS and gQA-S use 100-dim and 200-dim word embeddings respectively.
 Notice that gQA-S is comparable to HGN with respect to answer exact match and
answer $F_1$. Training gQA-S takes about 12hrs on a GTX 1080. We used a
90/5/5 split over 90400 examples found in the distractor training set. We did
not test on the dev set, nor did we submit to the HotPotQA leaderboards.

 \section{Discussion}

 We have presented a minimal extension to the original baseline of HotPotQA that
 performs comparably to SOTA with minimal preprocessing and minimal  hyper
 parameterization. The model presented here has the ability to easily
 transfer to single-doc QA, as the only difference would be the adjacency matrix.
 This might be explored in future work.
 Source code and pretrained models will be released upon publication.

A question remains about the performance of gQA-S when enriched with a large
language model, for instance RoBERTa, thus creating gQA-L. 

Furthermore an improved pooling mechanism, like multi-head attention, would
significantly aid in this direction. We have not included it here, since
our purpose was to investigate na\"{i}ve extension to the baseline model in order
to gauge the contribution of graph representation.

\begin{table}[t]
\centering
 \begin{tabular}{r | r | r | r | r}
 Model    & EM    & F$_1$ & Params& Base-LM\\
 gQA-XS   & 62.91 & 72.48 & 2.42M & None\\
 gQA-S    & \textbf{68.47} & 76.85 & 9.04M & None\\
 HGN      & 66.07 & 79.36 & $\geq$ 125M & RoBERTa-B\\
DFGN      & 56.31 & 69.69 & $\geq$ 110M & BERT-B\\
QFE       & 53.86 & 68.06 & N/A & None\\
KGNN      & 45.60 & 59.02 & N/A & None\\
Baseline & 44.60 & 59.02 & 960.1K & None\\
C2F-R & 67.98 & \textbf{81.24} & $\geq$ 667M & RoBERTa-L
 \end{tabular}
 \caption{Comparison of gQA to the baseline and others. At the time of
 our experiments, C2F Reader was SOTA.}
 \label{tab:my-table}
\end{table}
\bibliography{main}
\end{document}